\documentclass[conference]{IEEEtran}
\usepackage{blindtext, graphicx}
\usepackage{listings}
\usepackage{graphicx}
\usepackage[font=small]{caption}
\usepackage{xcolor,stfloats}
\usepackage[utf8]{inputenc}
\usepackage{amsmath}

\usepackage{bbding}
\usepackage{pifont}
\usepackage{wasysym}
\usepackage{amssymb}
\usepackage{multirow}

\usepackage{array} 
\renewcommand\arraystretch{1.3}

\title{A Single Multi-Task Deep Neural Network with a Multi-Scale Feature Aggregation Mechanism for Manipulation Relationship Reasoning in Robotic Grasping}

\author{
\IEEEauthorblockN{Mingshuai Dong}
\IEEEauthorblockA{Beijing University of Posts and Telecommunications, Beijing 100876, China\\ Email: dongmingshuai@bupt.edu.cn}
\IEEEauthorblockN{Yuxuan Bai}
\IEEEauthorblockA{Beijing University of Posts and Telecommunications, Beijing 100876, China\\ Email: 18522596667@bupt.edu.cn}
\IEEEauthorblockN{Shimin Wei}
\IEEEauthorblockA{Beijing University of Posts and Telecommunications, Beijing 100876, China\\ Email: wsmly@bupt.edu.cn}
\IEEEauthorblockN{Xiuli Yu*}
\IEEEauthorblockA{Beijing University of Posts and Telecommunications, Beijing 100876, China\\ Email: yxl@bupt.edu.cn}
}
\begin{document}

\maketitle
\thispagestyle{empty}
\pagestyle{empty}

%%%%%%%%%%%%%%%%%%%%%%%%%%%%%%%%%%%%%%%%%%%%%%%%%%%%%%%%%%%%%%%%%%%%%%%%%%%%%%%%
\begin{abstract}

Grasping specific objects in complex and irregularly stacked scenes is still challenging for robotics. Moreover, it is an important prerequisite for people to use robots to ensure the safety of the ontology and the environment during robot grasping. Therefore the robot is not only required to identify the object’s grasping posture but also to reason the manipulation relationship between the objects to improve the interaction safety. In this paper, we propose a manipulation relationship reasoning network with a multi-scale feature aggregation (MSFA) mechanism for robot grasping tasks. MSFA aggregates high-level semantic information and low-level spatial information in a cross-scale connection way to improve the environment understanding ability of the model. Furthermore, to improve the accuracy, we propose to use intersection features with rich location priors for manipulation relationship reasoning. Experiments are validated in VMRD datasets and real environments, respectively. Experimental results demonstrate that our method achieves state-of-the-art performance on the VMRD dataset. Furthermore, the experimental results in the real environment prove that our proposed method has the generalization ability and can be applied to actual scenarios.

Keywords: grasp detection, manipulation relationship, grasping order.

\end{abstract}

%%%%%%%%%%%%%%%%%%%%%%%%%%%%%%%%%%%%%%%%%%%%%%%%%%%%%%%%%%%%%%%%%%%%%%%%%%%%%%%%
\section{Introduction}

\begin{figure}
\centering
\includegraphics[width=\linewidth]{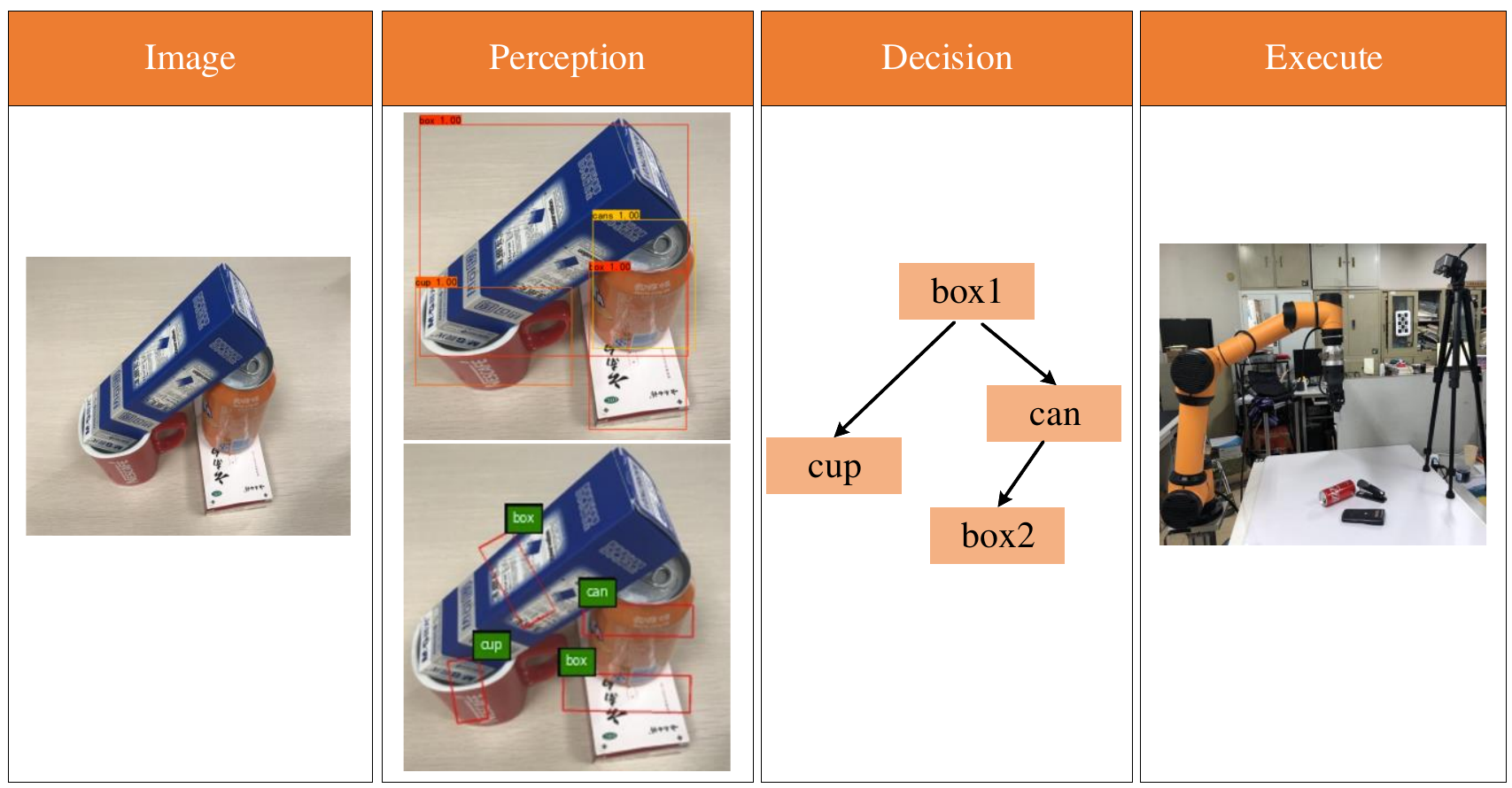}
\caption{Environment perception and decision-making are essential for robots to complete grasping operations in stacked scenes. It can recognize the objects in the environment, estimate the grasp position of each object and the position relationship between the objects, and make a reasonable grasp strategy.}
\label{fig1}
\end{figure}

With the development of robot technology, robotics can stably recognize and grasp objects in uncomplicated scenes\cite{2017A, 2, 3, redmon2015real}. However, grasping specific objects in cluttered or stacked complex scenes is always a challenging problem\cite{zhang2019multi}. As shown in Fig. \ref{fig1}, the necessary conditions for a robot to achieve safe grasping include perceiving objects and their grasping positions in the complex working scene, understanding the positional relationships between objects, and estimating a reasonable grasping strategy. Humans can quickly grasp target objects in any scene because humans can understand the environment and easily estimate the position relationship between objects and make reasonable decisions. Yet, this is a challenging task for robots. Therefore, it is essential to improve the environmental understanding ability of the robot so that it can reason about the position relationship between the objects in the stacked state and formulate a reasonable grasping strategy.

At present, researchers have achieved satisfactory results in the research of vision-based robot grasping in the single or multiple target objects scene. The structure of the above scene is straightforward, and there is no mutual occlusion and stacked position relationship between objects. Hence, the model only focuses on the grasping position of the object with low complexity. For complex stacked object scenes, \cite{2020Visual} proposed a multi-task visual manipulation relationship reasoning network(VMRN), which added the grasp detection task branch and a manipulation relationship reasoning branch between objects on the basis of the Faster-RCNN network. VMRN converts the manipulation relation reasoning task into a classification task. According to the target location estimated by the object detection branch, the features of the object pair on the common feature layer are intercepted to perform classification operations. \cite{2019A} proposes an end-to-end model based on the YOLO network, which can simultaneously realize object detection, grasp detection, and manipulation relationship inference. \cite{WOS:000724162300001} uses graph neural networks to model the features between object pairs and predict the manipulation relationship between objects. These methods have achieved satisfactory results.

Different from the above methods, we propose to further improve the model's performance by improving the model's ability to understand the environment. Firstly, we consider enriching the feature representation ability of the feature map to help the model understand the location relationship between different objects through richer fine-grained features and semantic features. In this paper, we propose a multi-scale feature aggregation module(MSFA) to fuse the information carried by features of different dimensions through two top-down and bottom-up feature fusions so as to enrich the representation ability of feature maps. Second, we utilize intersection features with rich location priors to assist the model in estimating the manipulation relationships between different objects. In general, if there is no intersection between two objects, there must be no occlusion and stacking position relations between objects. Therefore, this method is different from the complex relationship filtering network proposed by \cite{WOS:000724162300001}, and we improve the model's manipulation relationship discrimination ability through simple intersection features.

In summary, the main contributions of this paper are as follows:

A multi-task neural network based on multi-scale feature fusion is proposed to directly reason the object relationships and the manipulation order. 

In order to improve the efficiency and accuracy of the manipulation relationship inference network, we propose to use the intersection feature with location prior for inference, which has the relation filtering property to prune the objects that are unlikely to have a relationship.

The rest of this paper is organized as follows: Section II reviews the state-of-the-art in existing robotic manipulation relations. Section III details our proposed multi-task manipulation relationship inference network. In Section IV, we implement experiments to validate our proposed components as well as the performance of the model, and finally, in Section V, conclusions and discussions are drawn.

\section{Related work}
This section introduces state-of-the-art research on grasp detection and manipulation relationship reasoning in stacking scenarios.
\subsection{Robotic grasp detection}

Robotic grasping has always been the focus of researchers. In particular, vision-based deep learning technology is increasingly applied to robot grasping and achieves satisfactory results in specific scenarios. For example, \cite{2017A,2,3} transformed the grasp detection problem into a problem similar to object detection and achieved state-of-the-art results on the standard single-object Cornell grasp detection dataset. The above method uses a two-stage detection framework commonly used in object detection. In the first stage, candidate grasp configurations are generated. In the second stage, the offset between the grasp candidate and the ground truth is predicted. On the contrary, \cite{redmon2015real} proposed a one-stage model to use direct regression to predict the possible grasps in the input image. The direct regression method has a simple network structure but can only predict one grasp configuration per image. \cite{2020A} proposed a region-proposal-based grasp detection network, which first generates multiple reference anchors and then predicts the grasp configuration based on the features of these anchors. This method improves the perception ability of the one-stage model to the cluttered scene. \cite{20220511563781} proposed a two-stream convolutional neural network for simultaneous segmentation and grasp detection. This method uses a fully convolutional neural network to predict the grasp configuration for each pixel. \cite{2022When} and \cite{20223512677068} proposed introducing the Transformer architecture into the grasp detection model. Based on the powerful attention mechanism and global modeling ability of the Transformer architecture, the performance of the grasp detection model in the cluttered environment is improved.

The aforementioned methods have achieved satisfactory results in single or multi-objective environments. However, these models are difficult to apply to a random and complex stacking scenario. Therefore, applying these methods in industry or daily life is challenging. In addition, to expand the application scenarios of robots, it is necessary to have the ability to perceive various stacking scenes and reason about the spatial position relationship of objects. And according to the objects' spatial position relationship, the robot's manipulation sequence is further deduced.
\subsection{Manipulation relationship reasoning}
The reasoning of visual relations based on convolutional neural networks has achieved good performance \cite{tripathi2021using, dai2017detecting, liang2017deep}. However, the visual-based manipulation relationship reasoning in robotics is still rapidly developing.\cite{2020Visual} proposed to define the relationship that guides the orderly grasping of the robot as the manipulation relationship. The robot grasps objects correctly and efficiently in the complex stacked scene according to the manipulation relationship and does not cause damage to other targets. Moreover, \cite{2020Visual} proposed a visual manipulation relationship reasoning network, which first predicted the position relationship between objects in the stacked state and then constructed the robot manipulation relationship tree according to the position relationship. In practical applications, robots sequentially grasp objects in the working environment according to the manipulation relation tree. \cite{2019A} proposed a single-stage multi-task deep neural network that can simultaneously perform object detection, grasp detection, and object relation reasoning. Compared with the two-stage model of \cite{2020Visual}, the proposed method has higher computational efficiency. Different from [10] and \cite{2019A}, \cite{WOS:000724162300001} proposes to represent the positional relationship between objects in the form of a graph. And the graph network is introduced to predict the spatial position relationship of objects in the robot working scene. At the same time, in order to further reduce the amount of computation in the model reasoning stage, \cite{WOS:000724162300001} designed a relation filtering network to eliminate irrelevant object pairs before model reasoning.

According to the above studies, we can see that although there have been some recent works to solve the problem of visual manipulation relationship reasoning in an object-stacked environment, there are still challenges to improve the robot's ability to understand the environment and more accurately and effectively recognize the positional relationships of objects in the input scene.

\begin{figure*}
\centering
\includegraphics[width=\linewidth]{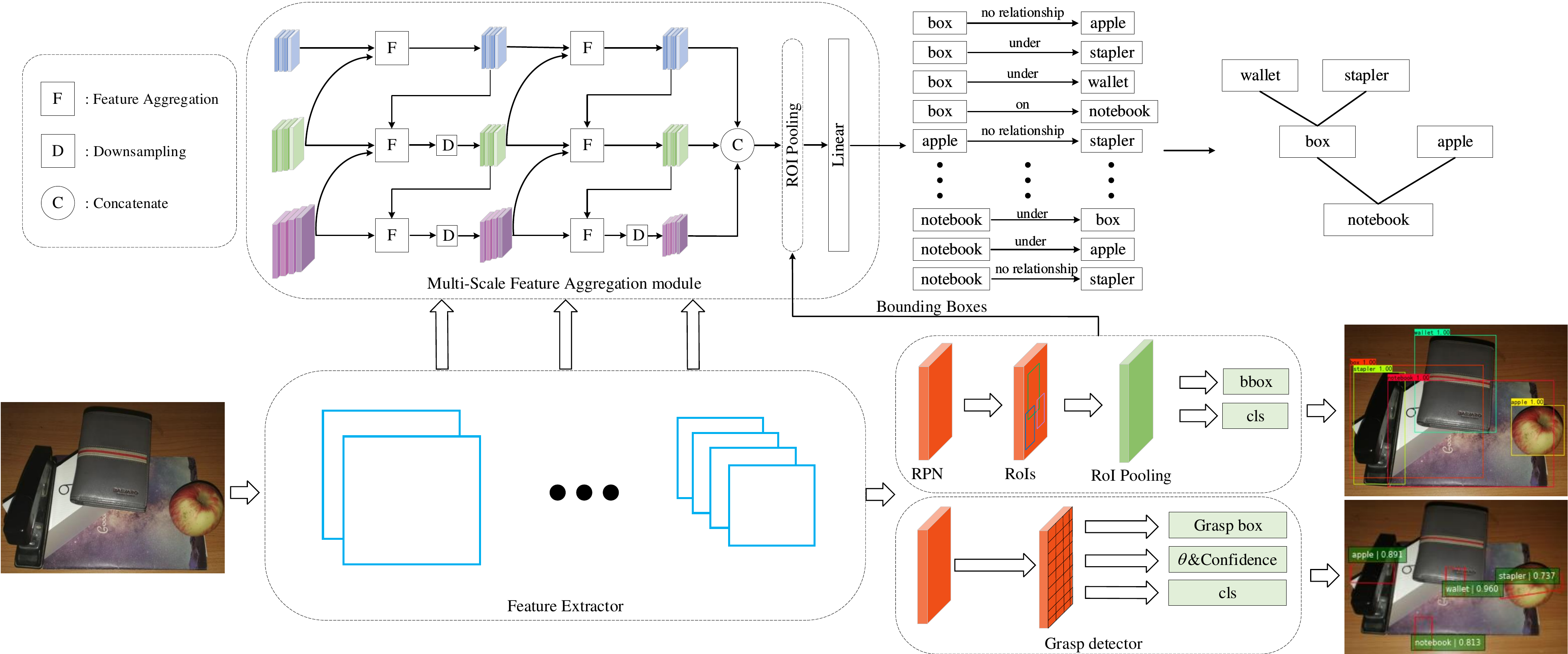}
\caption{The structure of our proposed model. It includes four parts: the backbone feature extractor, the grasp detector, the object detector, and the manipulation relationship reasoner based on the multi-scale feature aggregation module. Feature extractors usually use ResNet or VGG network for feature extraction and output feature maps. The object detector and the grasp detector are used to detect the object in the input image and the grasp position of the object, respectively. The manipulation relationship reasoning module uses multi-scale aggregated features to predict the positional relationship between objects and the manipulation order.}
\label{fig2}
\end{figure*}

\section{Method}
In this section, we first introduce the grasp representation and the representation of the relation representation methods used in this work. Then we give the details of the overall structure of our proposed model, including the detection and reasoning modules.

\subsection{Problem description}
\emph{Grasp Representation}: \cite{redmon2015real, 2013Deep} and other grasp detection works have proved that the 5D grasp configuration representation method $\{x,y,w,h,\theta\}$ can be widely used in robotic grasping tasks with parallel grippers. This representation is effective for robotic grasp scenarios that do not consider object categories. For ordered grasps or grasps with a specified object category, additional processing is needed to match the grasp representation to the correct object category. However, most of the current matching methods use post-processing, which reduces the running efficiency of the model. Different from the above methods, in this paper, we propose a 6D grasp representation $\{x,y,w,h,\theta,cls\}$ that adds category information to the original 5D grasp representation. Furthermore, the model uses the features of the grasp position to synchronously predict the category attribute of the grasp position in the prediction process. Thus, the structure of the model can be simplified, and the efficiency of the model can be improved.

\emph{Positional Relation Representation}: Constructing reasonable manipulation relationships in complex stacked scenes is necessary to guide the robot to grasp objects sequentially without causing potential damage to other objects. Inspired by \cite{2020Visual, 2019A, WOS:000724162300001}, \cite{2018A}, we also use the following three visual manipulation relations: object A is on object B (\emph{on}), object A is under object B (\emph{under}), and object A and object B have no relationships (\emph{no\_rel}). Through the manipulation relationship between the objects, we constructed the manipulation relationship tree to guide the robot to grasp the objects in the workspace safely and orderly.

\subsection{Architecture}
Fig. \ref{fig2} shows the architecture of our proposed model, which is a multi-task deep neural network with three parts: the backbone feature extractor, detection head, and reasoning head. In this network, we chose ResNet101 and VGG16 as the backbone feature extractor to extract the features of the RGB image. Then the detection and reasoning branches use the extracted features to perform tasks such as grasp detection, object detection, and manipulation relationship reasoning, respectively. To obtain satisfactory object detection accuracy for reasoning the manipulation relationships, we use the RPN structure in Faster R-CNN\cite{2017Faster} to extract object proposals. In the grasp detection branch, we use a single-stage prediction method to predict the grasp configuration according to the extracted features directly. For manipulation relationship reasoning, we propose a multi-scale feature aggregation mechanism to aggregate multi-dimensional features for manipulation relation prediction. The following is a detailed description of each part of our proposed model.

\emph{Object Detection}: We use Faster R-CNN architecture with higher accuracy for the object detection task. Therefore the object detection process consists of two stages. First, the RPN network estimates the region of interest(ROI) where the object may exist in the feature map and resize it to the same size through the pooling operation. Then, the object detector takes the pooled mini-batch ROIs features as input. The classifier and regressor without a hidden layer in the object detector are used to predict the position of the object bounding box and the classification result.

\emph{Grasp Detection}: Different from \cite{2020Visual}, we do not take the ROIs pooled features as the input of the grasp detector; we adopt the features extracted by the common feature extractor as the input of the grasp detector. We consider the axial symmetry of the ROI, which has limited filtering ability for background features. In addition, for the attribution of the grasp position, we use the direct prediction method to estimate the target object corresponding to the grasp position, which does not rely on the category information of the ROI.

In order to estimate the grasp position of the object in the robot working scene, we first divided the input feature map to the grasp detector into W*H grids. Inspired by our previous work \cite{20223512677068}, each pixel on the feature map corresponds to a grid. We design three anchors with different sizes at each grid point as priors. The grasp detector evaluates the offset and confidence of each prior anchor concerning the grasp position and predicts the object class corresponding to the grasp position. In addition, we adopt a classification approach to predict the rotation angle of the parallel gripper when the robot performs a grasp. We divided the grasp angle in the range (-90,90) into 19 categories equally. Therefore, in our work, the grasp detector head consists of three parts: a regression head for predicting the grasp position, a classification head for angle prediction and confidence evaluation, and a classification head for the predicted class.

\emph{Manipulation Relationship Reasoning}: In our work, we aggregate three different scales of features for robot manipulation relationship reasoning. For RGB images with an input of 600*600, we use 300*300, 100*100, and 40*40 feature maps for aggregating features at different scales. The aggregated features include high-resolution fine-grained features and high-dimensional features with rich semantic features. High-resolution fine-grained features can guide the model to learn more refined local features between different objects. Low resolution with high-dimensional features guides the model to pay more attention to global features and learn the spatial structure between objects in the input scene as a whole. At the same time, we perform two top-to-down and bottom-to-up feature fusions in this feature aggregation module.

To reason the manipulation relationship of the object in the robot working scene, we use the results of the object detector to crop the feature map and use the cropped features to make predictions. Different from the Object Pairing Pooling Layer proposed in \cite{zhang2019multi} to obtain the object pair features, we add the intersection features between object pairs on this basis. In contrast to union features, intersection features do not exist between all pairs of objects. Because when the objects are far away from each other, or there is no stacking relationship, the characteristic shown in the 2D RGB image is that there is no prominent intersection area between the two objects. Therefore, from this particular property of the intersection feature, we can conclude that the intersection feature can guide the model to filter out pairs of objects where no stacking relationship exists. Hence, the addition of intersection features can improve the discrimination ability of the model for manipulating relationships between objects.

After cropping the features using the object bounding box, we use ROI Pooling to adaptively resize the features into the same size. The features of object pairs we use include the features of object 1, the features of object 2, the union features, and the intersection features. It should be noted in particular that the object pair (O1, O2) is different from the object pair (O2, O1), and the manipulation relation between them cannot be interchanged entirely. As shown in Fig. \ref{fig2}, in the manipulation relationship reasoning stage, the pooled features are passed through the convolutional layer, and finally, the manipulation relationship of the object pair is classified through the fully connected network.

After obtaining the manipulation relations of each object pair in the robot working scene, we build the manipulation relations between objects in the whole scene as a relation tree, as shown in Fig. \ref{fig2}. When the robot grasps an object in the scene, it can obtain from the manipulation tree whether there is a leaf node of the object. If there is a leaf node, the robot should remove the leaf node first so as not to damage other objects.

\subsection{Loss function}
The loss function used by our network during training consists of three parts: object detection loss ${L_O}$, grasp detection loss ${L_G}$, and manipulation relation reasoning loss ${L_R}$.

The object detection loss ${L_O}$ includes Smooth L1 loss for supervised object location training and cross-entropy loss for supervised object category training, as shown in Eq.(\ref{Eq.1}) and Eq.(\ref{Eq.2}).

\begin{equation}\label{Eq.1}
\begin{aligned}
&Smooth\_L1(\{ gt\} ,\{ pre\} )\\
&= \frac{1}{n}\sum\limits_{i = 1}^n {\left\{ \begin{array}{l}
0.5 \times (g{t_i} - pr{e_i})_{}^2,{\rm{ }}if\left| {g{t_i} - pr{e_i}} \right| < 1\\
{\rm{ }}\left| {g{t_i} - pr{e_i}} \right| - 0.5,{\rm{  }}otherwise
\end{array} \right.}
\end{aligned}
\end{equation}

\begin{equation}\label{Eq.2}
H(gt,pre) =  - \sum\limits_{i = 1}^n {g{t_i}\log (pr{e_i})} 
\end{equation}

\begin{equation}\label{Eq.3}
    {L_O} = Smooth\_L1(\{ gt\} ,\{ pre\} ) + H(gt,pre)
\end{equation}

We apply BCE loss to evaluate the deviation between the ground truth and the predicted value in the grasp detection loss, as shown in Eq. (\ref{Eq.4}).

\begin{equation}\label{Eq.4}
    {L_G} =  - \sum\limits_{i = 1}^m {g{t_i}\log (pr{e_i})}  + (1 - g{t_i})\log (1 - pr{e_i})
\end{equation}

Where ${gt}$ is the ground truth of the object, ${pre}$ is the prediction result of the model, ${n}$ represents the number of objects in the image, and ${m}$ describes the number of grasp configurations in the input image.

In addition, we use a multi-class cross-entropy loss to supervise the training of the manipulated relation prediction branch.

\begin{equation}\label{Eq.5}
    {L_R} = \sum\limits_{(Ob{j_i},Ob{j_j})}^{} {\log (p_r^{(Ob{j_i},Ob{j_j})})}
\end{equation}
Where ${p_r^{(Ob{j_i},Ob{j_j})}}$ is the probability that the operation relation between (${Ob{j_1},Ob{j_2}}$) object pairs belongs to ${r}$, where $r \in \{ 1,2,3\} $ is the ground truth of the operation relation between the object pair (${Ob{j_1},Ob{j_2}}$).

In summary, in the process of network training, we define the whole loss of the model as ${L_{total}}$, as shown in Eq (\ref{Eq.6}).

\begin{equation}\label{Eq.6}
    {L_{total}} = {L_O} + \alpha {L_G} + \beta {L_R}
\end{equation}
In order to balance the training weights of the model, we set both $\alpha $ and $\beta $ to 5 during our experiments.

\section{Experiments set}
In this section, we mainly introduce the details of our experiments during training and testing, including the dataset used in the experiment process, the dataset's augmentation method, the experimental platform's details, and the model's evaluation metrics.

\subsection{Dataset}
We used the VMRD dataset proposed in \cite{2018Visual} to train the model. In this dataset, each image contains multiple objects, and there are occlusions and stacks between objects. The dataset consists of 4683 images, where objects in each image are accurately labeled with location, category, grasp configuration, and manipulation relationships between objects. The training set and test set are set to 4233 and 450 images, respectively, in this dataset.

During the training process, we take advantage of online random data augmentation to prevent overfitting and improve the generalization ability of the model for different working scenarios. Data augmentation methods include image scaling, flipping, and gamut transformation. Hence, the images fed into the model in each training iteration are random and different. During model testing, we do not augment the input image but only resize the image to a fixed size.
\subsection{Implementation details}
We use Faster R-CNN to train the object detection module, where the backbone feature extractor adopts the pre-trained VGG16 and Resnet101, respectively. In addition, the two-stage method is used to train our network. We first train the feature extraction and object detection modules and then fix the training parameters to train the manipulation relation reasoning and grasp detection modules synchronously. 

Our model is implemented using Pytorch deep learning model framework. And our models are trained on GTX 2080TI with 11G memory. During model training, we set the batch size to 8 and the learning rate to 0.001, which is divided by 10 every 10,000 iterations. Moreover, the object detection module is trained for 100 epochs separately, and the grasp detection and manipulation relation reasoning modules are trained together for 200 epochs.

\subsection{Metrics}
In our model, we evaluate the performance of each task separately using different methods. For the object detection task, we use mAP to measure the detection results of all classes of objects. In the grasp detection task, we use the rectangle metric to validate the grasp detection results of the model. The rectangle metric contains two parts, a) the Jacquard index between the predicted grasping rectangle and the ground truth is greater than 25\%; b) The difference between the angle of the predicted grasping rectangle and the ground truth is within 30°. We consider a predicted as correct when the above two metrics are simultaneously satisfied between the predicted grasp position and the ground truth.

In order to verify the performance of the manipulation relation reasoning branch, inspired by \cite{WOS:000724162300001}, we also use the following three metrics:

\textbf{OR} (Object-based Recall): This metric is used to evaluate the recall of the model prediction results, which we consider correct when both the category of a pair of objects as well as the manipulation relationship between them are predicted correctly.

\textbf{OP} (Object-based Precision): The average precision of the three manipulation relationships between object pairs predicted by the model.

\textbf{IA} (Image-based Accuracy): This metric is the evaluation accuracy based on all objects in the image. The prediction of an image is considered correct when all object categories in the image are predicted correctly and all manipulation relationships between object pairs are predicted correctly.

\section{Results and analysis}
In this section, we first describe the details and results of ablation experiments to demonstrate the effectiveness of our proposed method. We then set up contrast experiments to compare our results with state-of-the-art methods. Finally, we design physical grasping experiments in the stacking scene to verify the performance of our model.

\subsection{Ablation study}

\begin{table}[]
\caption{Self-comparison experiments on the VMRD dataset under different conditions.}
\begin{center}
\scalebox{1.2}{
\begin{tabular}{ccccc}
\hline
  & \multirow{1}{*}{\textbf{MSFA}} & \multirow{1}{*}{\textbf{Intersection feature}} & \multirow{1}{*}{\textbf{OR (\%)}}& \multirow{1}{*}{\textbf{IA (\%)}} \\  \hline
A & ×    & ×                    &   87.3     &  68.4 \\ 
B & \checkmark    & ×                    & 90   & 73.8  \\ 
C & ×    & \checkmark                    & 88.2   &  69  \\ 
D & \checkmark    & \checkmark                    &90.5  & 74.7    \\ \hline
\end{tabular}
}
\end{center}
\end{table}

\begin{figure*}
\centering
\includegraphics[width=\linewidth]{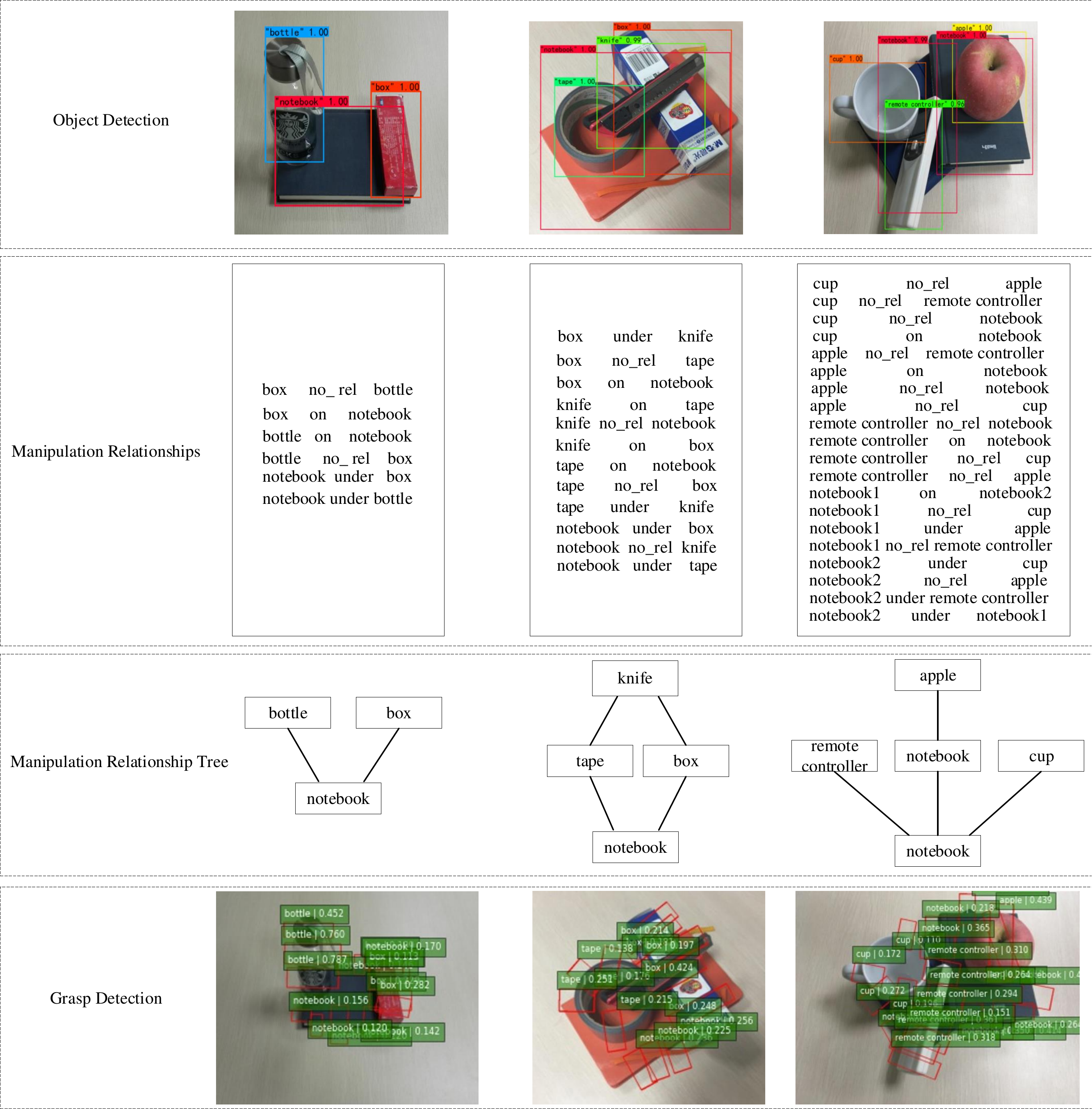}
\caption{Experiment results of our proposed model on the VMRD dataset. The first row is the object detection results of the model in scenes with different numbers of objects. The second row is the prediction result of the operation relationship. The third is the robot manipulation relation tree estimated by the inference result of the manipulation relationship. The fourth row shows the grasp detection results.}
\label{fig3}
\end{figure*}

Table I summarizes the results of our ablation studies on the VMRD test dataset. We utilize the VGG backbone to verify the effectiveness of our proposed improvements through a series of self-contrast experiments. In experiment A, the multi-scale feature aggregation (MSFA) module and intersection features were not used for manipulation relationship reasoning. In experiment B, the MSFA module was added to verify the contribution of multi-scale features to the model performance. Experiment C adds the intersection features of object pairs to the inference of manipulation relations based on Experiment A. Finally, experiment D shows the performance of our proposed model.

Combining the results of Experiment A and Experiment B, we can find that after the model uses the MSFA module, the OR index is improved by 2.7\%, and the IA index is improved by 5.4\%. We consider that the MSFA module aggregates low-dimensional features and high-dimensional features with richer fine-grained features compared to the feature map used by the original model to manipulate relationship reasoning. Thus, the aggregated feature map has a more powerful feature representation ability, and it can better learn the features between object pairs, thereby improving the model's performance.

Comparing Experiments A and C, we can see that adding intersection features to the manipulation relation inference process leads to a slight rise in the performance of the model on the OR and IA metrics. The reason for this phenomenon is that intersection features can help the model improve the recognition accuracy of the manipulation relationship between no-relationship (\emph{no\_rel}) object pairs. Nevertheless, there is no apparent contribution to reasoning about other manipulation relationships. However, it is also sufficient to demonstrate that the positional priors between objects in the intersection features contribute to the reasoning about the manipulation relationships.

Experiment D shows the experimental results of our proposed model, which uses both the MSFA module and the intersection features. Compared with experiment A, we can find that the OR index of the model is improved by 3.2\%, and the IA index is improved by 6.3\% after adding the MSFA module and intersection features. This demonstrates that our proposed method improves the manipulation relation inference performance of the model.

\subsection{Results for manipulating relationship reasoning}

\begin{table*}[]
\caption{Performance summary of different manipulation relationship reasoning methods on the VMRD dataset.}
\begin{center}
\setlength{\tabcolsep}{7mm}{
\begin{tabular}{ccccccc}
\hline
\multirow{2}{*}{\textbf{Backbone}}       & \multirow{2}{*}{\textbf{Model}}      & \multirow{2}{*}{\textbf{mAP(\%)}}       & \multirow{2}{*}{\textbf{OR(\%)}}       &\multirow{2}{*}{ \textbf{OP(\%)}}      & \multirow{2}{*}{\textbf{IA(\%)}}        & \multirow{2}{*}{\textbf{Time (ms)}}   \\ 
\\\hline
\multirow{5}{*}{ResNet101} & Multi-task CNN\cite{zhang2019multi} & -             & 86.0          & \textbf{88.8} & 67.1          & -           \\  
                           & VMRN\cite{2020Visual}           & 95.4          & 85.4          & 85.5          & 65.8          & 98          \\  
                           & GVMRN\cite{WOS:000724162300001}          & 94.5          & 86.3          & 87.1          & 68.0          & 102         \\  
                           & GVMRN-RF\cite{WOS:000724162300001}       & 94.6          & 87.4          & 87.9          & \textbf{69.3} & \textbf{67} \\  
                           & \textbf{Ours}  & \textbf{96.3} & \textbf{87.7} & 88.7          & 67.1          & 110         \\ \hline
\multirow{4}{*}{VGG16}     & VMRN\cite{2020Visual}           & 94.2          & 86.3          & 88.8          & 68.4          & 71          \\  
                           & GVMRN\cite{WOS:000724162300001}          & 95.4          & 87.3          & 89.6          & 69.7          & 92          \\  
                           & GVMRN-RF\cite{WOS:000724162300001}       & 95.4          & 89.1          & 89.7          & 70.9          & \textbf{58} \\  
                           & \textbf{Ours}  & \textbf{97.4} & \textbf{90.0} & \textbf{89.8} & \textbf{73.8} & 90          \\ \hline
\end{tabular}
}
\end{center}
\end{table*}

\begin{table*}[]
\caption{Summary of performance of different models in scenes with different numbers of objects.}
\begin{center}
\setlength{\tabcolsep}{7mm}{
\begin{tabular}{ccccccc}
\hline
\multirow{3}{*}{\textbf{Backbone}}  & \multirow{3}{*}{\textbf{Model}} & \multicolumn{5}{c}{\textbf{Image-based accuracy (IA)}}                                                                                                                                  \\ \cline{3-7} 
                           &                        & \multicolumn{1}{c}{\multirow{2}{*}{\textbf{Total (\%)}}} & \multicolumn{4}{c}{\textbf{Object number per image}}                                                                                 \\ \cline{4-7} 
                           &                        & \multicolumn{1}{c}{}                            & \multicolumn{1}{c}{\textbf{2}}             & \multicolumn{1}{c}{\textbf{3}}             & \multicolumn{1}{c}{\textbf{4}}             & \textbf{5}             \\ \hline
\multirow{5}{*}{ResNet101} & Multi-task CNN\cite{zhang2019multi}         & \multicolumn{1}{c}{67.1}                        & \multicolumn{1}{c}{87.7}          & \multicolumn{1}{c}{64.1}          & \multicolumn{1}{c}{56.6}          & 72.9          \\  
                           & VMRN\cite{2020Visual}                   & \multicolumn{1}{c}{65.8}                        & \multicolumn{1}{c}{-}             & \multicolumn{1}{c}{-}             & \multicolumn{1}{c}{-}             & -             \\  
                           & GVMRN\cite{WOS:000724162300001}                  & \multicolumn{1}{c}{68.0}                        & \multicolumn{1}{c}{90.0}          & \multicolumn{1}{c}{68.8}          & \multicolumn{1}{c}{60.3}          & 56.2          \\  
                           & GVMRN-RF\cite{WOS:000724162300001}               & \multicolumn{1}{c}{69.3}                        & \multicolumn{1}{c}{\textbf{91.4}}          & \multicolumn{1}{c}{\textbf{69.5}}          & \multicolumn{1}{c}{62.1}          & 58.9          \\  
                           & \textbf{Ours}          & \multicolumn{1}{c}{\textbf{73.0}}                            & \multicolumn{1}{c}{82.6}              & \multicolumn{1}{c}{68.1}              & \multicolumn{1}{c}{\textbf{70.1}}              &   \textbf{71.4}            \\ \hline
\multirow{4}{*}{VGG16}     & VMRN\cite{2020Visual}                   & \multicolumn{1}{c}{68.4}                        & \multicolumn{1}{c}{-}             & \multicolumn{1}{c}{-}             & \multicolumn{1}{c}{-}             & -             \\  
                           & GVMRN\cite{WOS:000724162300001}                  & \multicolumn{1}{c}{69.7}                        & \multicolumn{1}{c}{91.4}          & \multicolumn{1}{c}{69.9}          & \multicolumn{1}{c}{62.9}          & {58.9}          \\  
                           & GVMRN-RF\cite{WOS:000724162300001}               & \multicolumn{1}{c}{70.9}                        & \multicolumn{1}{c}{\textbf{92.9}}          & \multicolumn{1}{c}{70.7}          & \multicolumn{1}{c}{64.6}          & 61.6          \\  
                           & \textbf{Ours}          & \multicolumn{1}{c}{\textbf{73.8}}               & \multicolumn{1}{c}{83.1} & \multicolumn{1}{c}{\textbf{68.4}} & \multicolumn{1}{c}{\textbf{70.6}} & \textbf{73.1} \\ \hline
\end{tabular}
}
\end{center}
\end{table*}

Table II presents the results of different manipulation relationship reasoning models. We can see that our method has better performance. Compared with the method proposed in \cite{zhang2019multi}, our method has comparable results. Especially for the OR metric, the accuracy of our proposed method is 1.7\% higher. Compared to VMRN\cite{2020Visual}, which has a similar network structure, our method achieves better performance in all indicators in two networks with different backbone structures. We can conclude that the method based on multi-scale feature aggregation can better learn the location features between objects in the image to infer the manipulation relationship between objects more accurately.

Compared with GVMRN and GVMRN-RF \cite{WOS:000724162300001} using graph networks, our method outperforms the above models in terms of OR and OP metrics in the ResNet101 backbone model. In the model with a VGG16 backbone, our method is better than GVMRN and GVMRN-RF in OR, OP, IA, and mAP. This also further demonstrates the effectiveness of our proposed feature aggregation method and the involvement of intersection features in manipulation relationship reasoning. The running time of our proposed model has a slight gap from that of VMRN and GVMRN-RF, but it does not affect the real-time running effect of the robot.

Table III presents the reasoning results of the manipulation relationships for the different numbers of stacked objects. When there are only two objects in the robot working scene, the accuracy of our model manipulation relationship inference is only 83.1\%, which is lower than GVMRN and GVMRN-RF proposed by \cite{WOS:000724162300001}. When the number of objects in the robot working scene is more fantastic than four, the accuracy of our model on the test set is much better than that of GVMRN and GVMRN-RF. This trend of accuracy change is similar to the Multi-task CNN model proposed by \cite{zhang2019multi}, and the reason for this phenomenon is that we use a similar network structure and the same test set in \cite{zhang2019multi}. Compared with the Multi-task CNN method, we can conclude that our proposed method dramatically improves the inference accuracy of manipulation relations, especially in scenes with more than three multi-objects.

The three examples in Fig. \ref{fig3} demonstrate the output of our proposed reasoning network for visual manipulation relations. From the experimental results, it can be seen that our proposed model can accurately predict the manipulation relationship between objects in different numbers of object scenes. And according to the manipulation relationship, the manipulation relationship tree can be constructed to guide the robotic grasping. Furthermore, the last row shows the grasping positions predicted by the model for different objects.
\subsection{Robot grasping experiment}

\begin{figure}
\centering
\includegraphics[width=\linewidth]{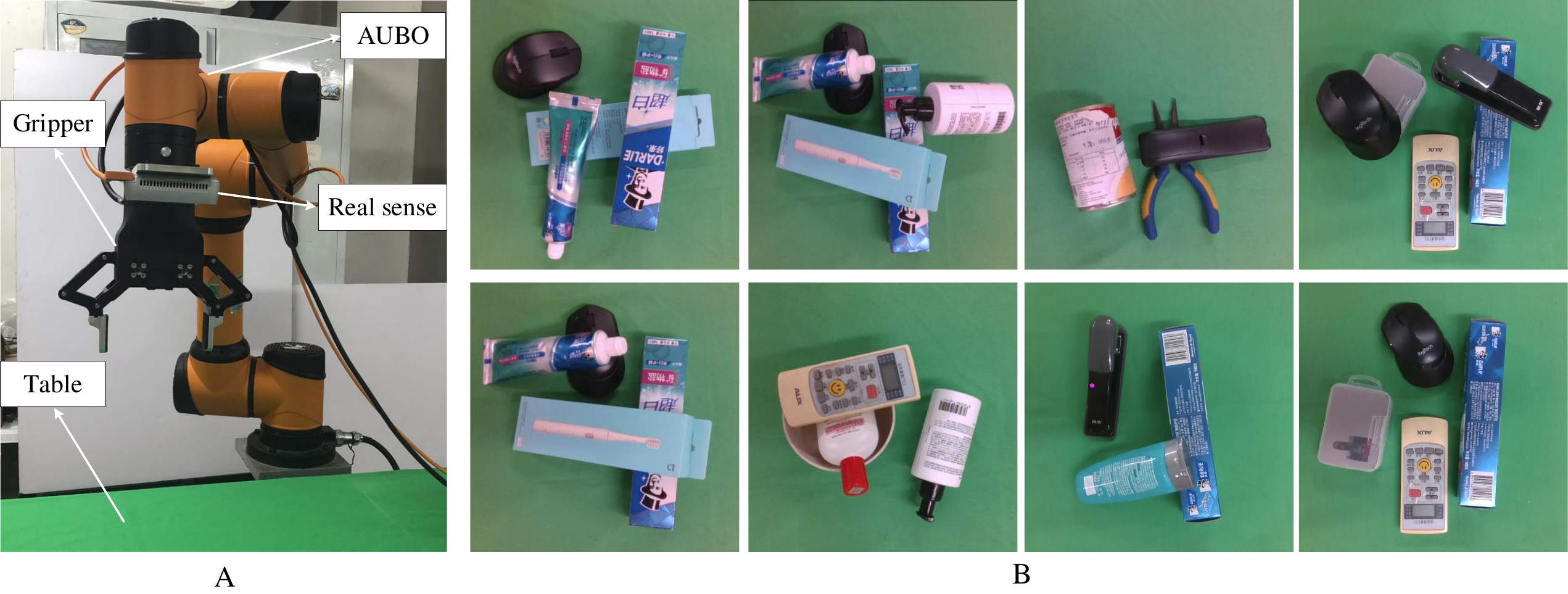}
\caption{Experimental environment for robot grasping. (A) the physical platform used for robotic grasping experiments, and (B) some examples of grasping experiment scenarios.}
\label{fig4}
\end{figure}

To verify the performance of our proposed model in real scenarios, we deploy the model on a real grasping platform. The overall layout of the platform is shown in Fig. \ref{fig4}(A). The experimental platform contains a 6-DOF AUBO-i5 robotic arm, a RealSense-D435i depth camera, and a parallel plate gripper. Fig. \ref{fig4}(B) illustrates a portion of the scene during the real experiment of the robot. Each scene contains three to five randomly stacked placed objects, which are used to verify the performance of our model in real scenes as well as the generalization ability.

\begin{figure}
\centering
\includegraphics[width=\linewidth]{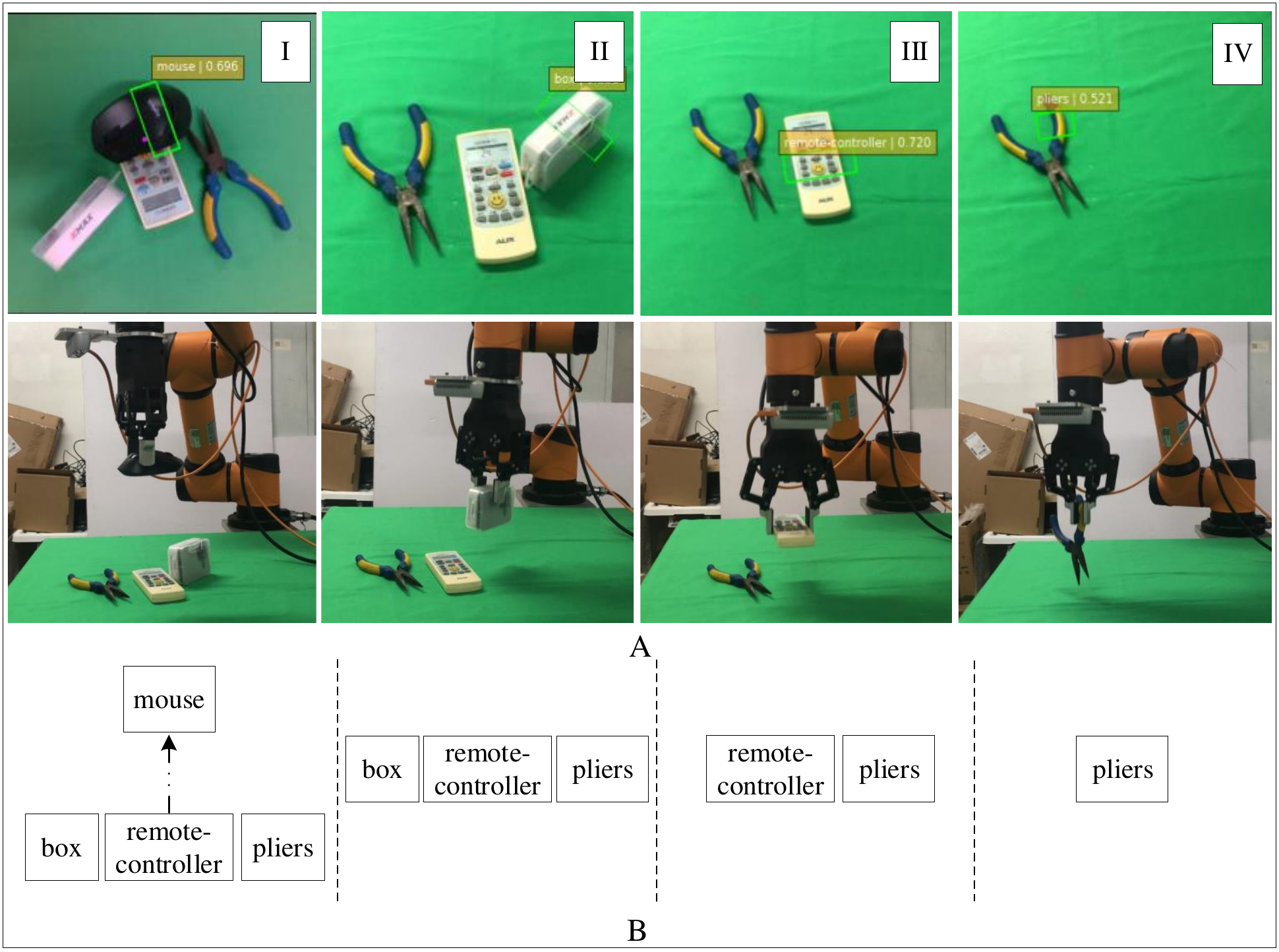}
\caption{The sorting process of the robot in a stacking scenario.}
\label{fig5}
\end{figure}

\begin{figure}
\centering
\includegraphics[width=\linewidth]{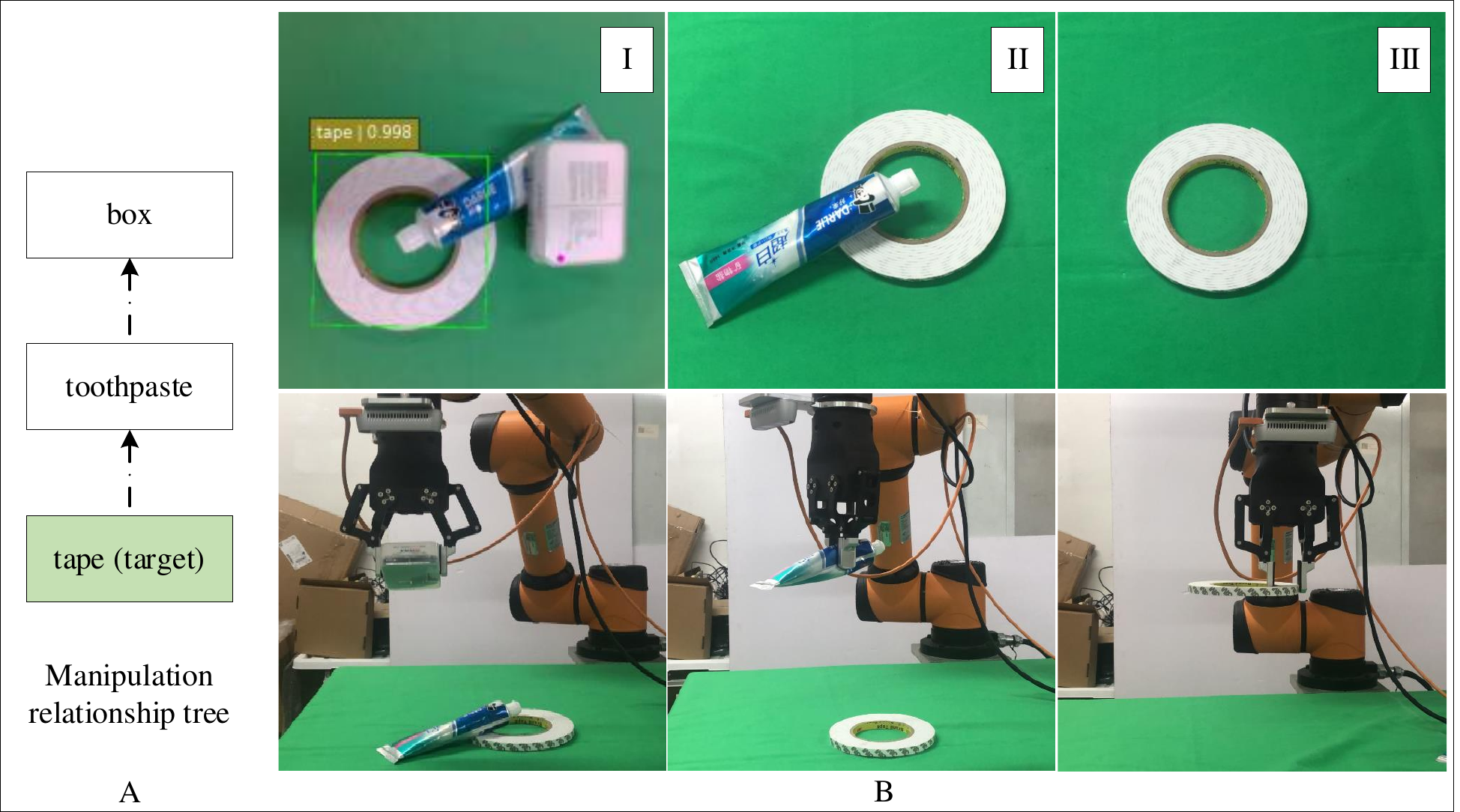}
\caption{Robotic grasping process with "tape" as the target.}
\label{fig6}
\end{figure}

As shown in Fig. \ref{fig5}, the sorting process of the robot in the multi-object stacking scene is constrained by the spatial position relationship and category between objects. The robot detects objects in the working scene and deduces the positional relationship between each object to generate a manipulation relationship tree. The robot consecutively grasps from top to bottom according to the recommendation of the manipulation relation tree.

For the scenario presented in Fig. \ref{fig6}, the grasping target of the robot is the tape. However, we found that the tape was pressed under the toothpaste, and a box was pressed on top of the toothpaste. Therefore, if the robot grasps the tape directly, then the toothpaste and the box will be displaced indeterminately. If the toothpaste and the box are fragile objects, then this blind operation will damage the above objects. Hence, our model reasoned that the sequence of actions between objects in this scene was to grasp the box first, then the toothpaste, and finally the tape, as shown in Fig. \ref{fig6}.  Through the above experiments, we can conclude that our model can infer the position relationship between objects in the multi-object stacked scene, and can guide the robot to grasp the target object in a reasonable manipulation sequence.

\section{Conclusion}

In this paper, we propose a multi-task model for multi-object stacked scenes to reason about the positional relationships between objects, formulate grasping plans, and estimate the grasping positions of different objects. Thus, the robot is guided to achieve autonomous grasping. In addition, in order to improve the model's ability to understand the location relationship between objects, we propose a multi-scale feature aggregation network to enrich the representation ability of the model and introduce intersection features to increase the location relationship prior. Compared with the existing models, our method improves the reasoning accuracy of the model by 4.5\%, which fully proves the effectiveness of our method. Moreover, we deploy our proposed model in a physical experimental scenario to further verified the usability of the model.

\addtolength{\textheight}{-12cm}   % This command serves to balance the column lengths

%%%%%%%%%%%%%%%%%%%%%%%%%%%%%%%%%%%%%%%%%%%%%%%%%%%%%%%%%%%%%%%%%%%%%%%%%%%%%%%%

\bibliographystyle{IEEEtran}  
\bibliography{referencr}

\end{document}